\documentclass{article}

  \usepackage[preprint]{neurips_2026}

\usepackage[utf8]{inputenc} 
\usepackage[T1]{fontenc}    
\usepackage{hyperref}       
\usepackage{url}            
\usepackage{booktabs}       
\usepackage{amsfonts}       
\usepackage{amsmath}
\usepackage{nicefrac}       
\usepackage{microtype}      
\usepackage{xcolor}         
\usepackage{subcaption}
\usepackage{wrapfig}
\usepackage{gensymb}
\usepackage{fontawesome5}

\hypersetup{
    colorlinks=true,
    linkcolor=black,
    filecolor=magenta,      
    urlcolor=black,
    citecolor=blue,
    pdftitle={Your Paper Title},
}

\newcommand{\hficon}{\raisebox{-0.7ex}{\includegraphics[height=1.3em]{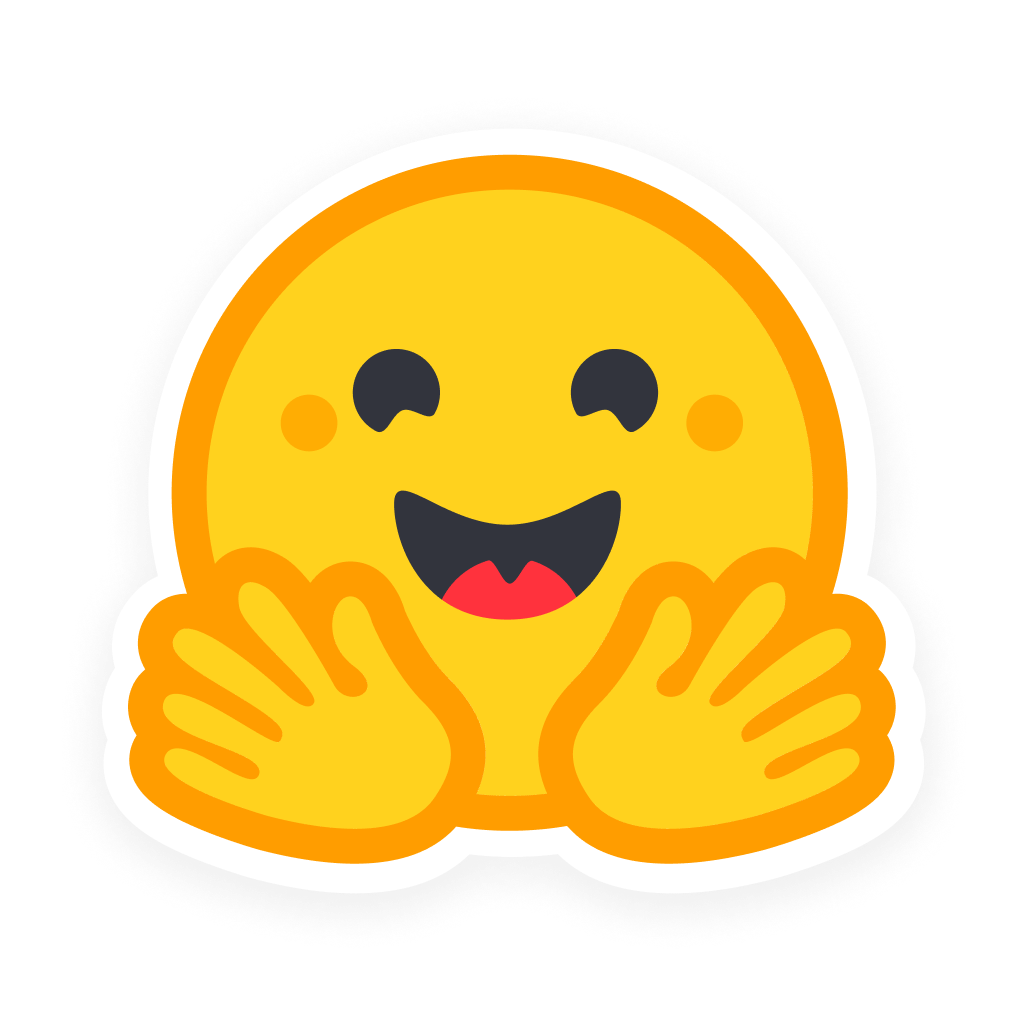}}}

\usepackage[disable]{todonotes}
\title{\texttt{Kine2Go}: \\ Kinematic dataset for the Unitree Go2 robot with diverse gaits and motions}

\author{
    Władysław Pałucki \\
    University of Warsaw\\
    \texttt{w.palucki@uw.edu.pl} \\
    \And
    Paweł Siwak \\
    University of Warsaw\\
    \texttt{pl.siwak@student.uw.edu.pl} \\
    \AND
    Krzysztof Ciebiera\\
    University of Warsaw\\
    \texttt{k.ciebiera@uw.edu.pl} \\
    \And
    Marek Cygan\\
    University of Warsaw\\
    \texttt{cygan@mimuw.edu.pl} \\
}

\begin{document}

\maketitle

\begin{abstract}
    The recent popularity of robotics, combined with the steadily decreasing cost of robotic hardware, has lowered the entry barrier to robotics research and enabled rapid advancements in the field. One of the primary examples is the Unitree Go2 quadruped robot, which is often used by researchers in the areas of locomotion, navigation, control, and others. Many researchers use the Go2 robot in combination with techniques like imitation learning, reinforcement learning, and behavioral cloning to allow machine learning systems to take full control of the robot. At the same time, many of those techniques require demonstration data consisting of the robot's kinematics information and actions applied to the motors. Obtaining such data is difficult, requires building complex pipelines, and can take significant time. To aid in those kinds of efforts, we present \texttt{Kine2Go} - a dataset with 800 diverse gait kinematics trajectory motion data for the Unitree Go2 robot, derived from 40 distinct policies. Our pipeline accepts data from various quadruped morphologies and translates them to a Go2-compatible format. Then we use Reinforcement Learning to train policies following a given motion, and finally we gather data from those policies, which grants robust, perturbed kinematic data with corresponding motor-level actions. 
\end{abstract}

\vspace{0.2cm} 
\begin{center}
    \href{https://github.com/nomagiclab/kine2go-pipeline}{\faGithub \hspace{0.35em} \texttt{Code}}
    \quad \quad \textcolor{gray}{|} \quad \quad 
    \href{https://huggingface.co/datasets/MIMUW-Robotics/kine2go}{\hficon \hspace{0.35em} \texttt{Dataset}}
\end{center}
\vspace{0.2cm} 

\section{Introduction} \label{sec:intro}

The newest generation of robots can achieve unprecedented speed and agility while moving, even through complex terrain. Most of these feats, however, are achieved through low-level control techniques like PIDs, MPC, etc. On the other hand, machine learning frameworks that promise to learn end-to-end control, like Reinforcement Learning (RL), usually struggle to produce movements that look \textit{reasonable}. Without proper regularization, RL tends to produce policies that use extreme actions and unstable, erratic gaits. For most real-world robotics, it is desirable to produce motions that are smooth, natural, symmetric, and animal-like.

Depending on the exact method, regularizing policies can be achieved through many different techniques. In Reinforcement Learning, it is possible to design a reward that will produce a motion of desired properties. However, designing those rewards, often called "reward shaping", is a hard, tedious task requiring a high degree of trial and error. At the same time, RL has a tendency to try to bypass or hack the rewards, which makes the problem even more difficult \citep{reward_hacking}.

The most straightforward approach is motion imitation \citep{motionimitation} or behavioral cloning, where a policy is trained to mimic an expert's trajectories. Those methods tend to yield desirable results, however, they require an expert dataset of motions. In most cases, they are also very specialized, meaning that a single policy is able to recreate a single type of motion or trajectory. While it can be enough for some applications, in most cases, it would be best to produce a singular policy that can, depending on the situation and command, alternate between different types of gaits and skills, and is not limited to a particular style of motion.

Finally, there are some approaches \citep{metamotivo, bfmzero, nvidiasonic, yang2025generalizedanimalimitatoragile} that use a whole dataset of diverse motions as a regularizing factor for a policy. Building on top of them, they create a conditioned policy able to perform multiple different motions depending on the input. Those kinds of efforts are particularly interesting because they can lead to more general, foundational policies able to perform a wide range of tasks. 

We present \texttt{Kine2Go} - a kinematic dataset with over 800 trajectories capturing complex, diverse gaits of the Unitree Go2 robot. We hope that our dataset will aid particularly in the last of the aforementioned approaches, where a diverse dataset is the necessary basis. By utilizing multiple data sources and coalescing them into a single pipeline, we were able to produce a comprehensive gait collection, covering many possible movements. The dataset is collected as kinematic motion for a robot, including position, rotation, velocity, angular velocity, joint position, and joint velocities, as well as actions (motor usage) taken by the policy to achieve each trajectory. Through conditioning various emerging methods on such a dataset, it could be possible to create a policy capable of natural and diverse locomotion, opening new directions in research on robotics, both in simulation and the real world. Such approaches described above were possible in the humanoid space due to the large amount of Motion Capture data. Quadrupedal morphology lacks similar scale datasets and, as such, has not seen similar efforts. Our work aims to address this gap by introducing a large, diverse, and novel dataset.

In addition to the dataset, we also contribute a pipeline that can be easily extended and reused to either gather even more trajectories for the Go2 robot, from new data sources, or, after minor modifications, transplant our approach to another morphology. Crucially, our pipeline uses the Genesis \citep{Genesis} engine, which is fast, easy to work with, and well optimized for consumer-grade GPU hardware. This can further aid researchers willing to experiment on different robots and bring their own data sources.

\section{Related Work} \label{sec:related_work}

\subsection{Large-scale robot learning datasets}

Large-scale datasets have become a central mechanism for improving generalization and reproducibility in robot learning. Early multi-robot resources such as RoboNet aggregated interaction data across several robot platforms and showed that pretraining on shared robot experience can improve adaptation to held-out platforms and environments \citep{DBLP:journals/corr/abs-1910-11215}.

More recent manipulation datasets, including BridgeData V2, DROID, and Open X-Embodiment, further demonstrate the value of collecting diverse demonstrations across tasks, objects, environments, and hardware embodiments
\citep{walke2024bridgedatav2datasetrobot, khazatsky2025droidlargescaleinthewildrobot, embodimentcollaboration2025openxembodimentroboticlearning}. 
These datasets are primarily oriented toward visual manipulation and high-level embodied behavior. In contrast, \texttt{Kine2Go} focuses on low-level legged locomotion: it provides Go2-compatible kinematic trajectories together with the motor-level actions used to realize them in simulation.

This distinction is important because locomotion policies operate at the motion-control layer. For a quadruped, a policy does not only need to know the task objective or visual scene; it must infer actuator commands from the robot's current dynamic state, including body orientation, velocities, joint configuration, and recent control history. This is why prior motion-imitation datasets and systems separate kinematic reference motions from state-action rollouts: reference motions describe the desired behavior, while rollouts provide the control information needed to reproduce it in simulation or on hardware \citep{Peng_2018}. \texttt{Kine2Go} therefore complements visual robot datasets by providing Go2-specific kinematic and action trajectories for low-level locomotion learning.

\subsection{Kinematic motion datasets}

Large-scale motion datasets have become an important substrate for learning motion models and controllers. In human motion learning, datasets such as Human3.6M provide 3D joint trajectories and whole-body motion capture data for pose estimation, motion analysis, and transfer to robot embodiments \citep{zhu2023h3wbhuman36m3dwholebody}. AMASS further standardized heterogeneous human mocap collections by converting them into a common SMPL body representation, creating a unified motion corpus that has since been used for motion generation, imitation learning, and control \citep{mahmood2019amassarchivemotioncapture}. This standardization principle is central to our work: heterogeneous motions become substantially more reusable once they are expressed in a shared embodiment-specific representation.

Comparable resources for animal and quadruped motion are more limited.
Animal3D provides 3D pose and shape annotations for diverse mammal species using SMAL parameters, while PFERD provides dense marker-based 3D motion capture for horse locomotion \citep{xu2024animal3dcomprehensivedataset3d,
li2024poses}. These datasets are valuable sources of animal morphology and kinematics, but they are not directly expressed in the morphology, state space, or action space of a quadruped robot such as the Unitree Go2. \texttt{Kine2Go} addresses this gap by converting heterogeneous quadruped motions into a Go2-compatible representation and augmenting the resulting trajectories with policy-generated motor actions.

\subsection{Motion-control and quadruped datasets}

Kinematic motion data describes what motion should be reproduced, but it does not directly provide the motor commands required to realize that motion under robot dynamics. Motion-imitation pipelines, therefore, commonly retarget a reference motion to the target morphology, train a controller to track it, and use the resulting simulated rollouts as control data. This strategy has been used for agile quadruped locomotion, where animal reference motions were retargeted and imitated by a Laikago robot \citep{yu2021visuallocomotion}.

The closest dataset-level analogue to our setting is MoCapAct, which converts human mocap clips into humanoid rollout datasets containing proprioceptive observations and actions \citep{wagener2023mocapactmultitaskdatasetsimulated}.

\texttt{Kine2Go} follows the same principle for quadruped locomotion, but targets the Unitree Go2 morphology.
Rather than releasing only retargeted reference trajectories, \texttt{Kine2Go} provides Go2-compatible rollouts containing root state, joint state, velocities, and policy-generated motor actions. This makes the dataset directly applicable to behavioral cloning, offline control, motion-conditioned policy learning, and evaluation of Go2 locomotion models.

\section{Dataset gathering pipeline} \label{sec:pipeline}
\begin{figure}
  \begin{center}
    \includegraphics[width=0.95\textwidth]{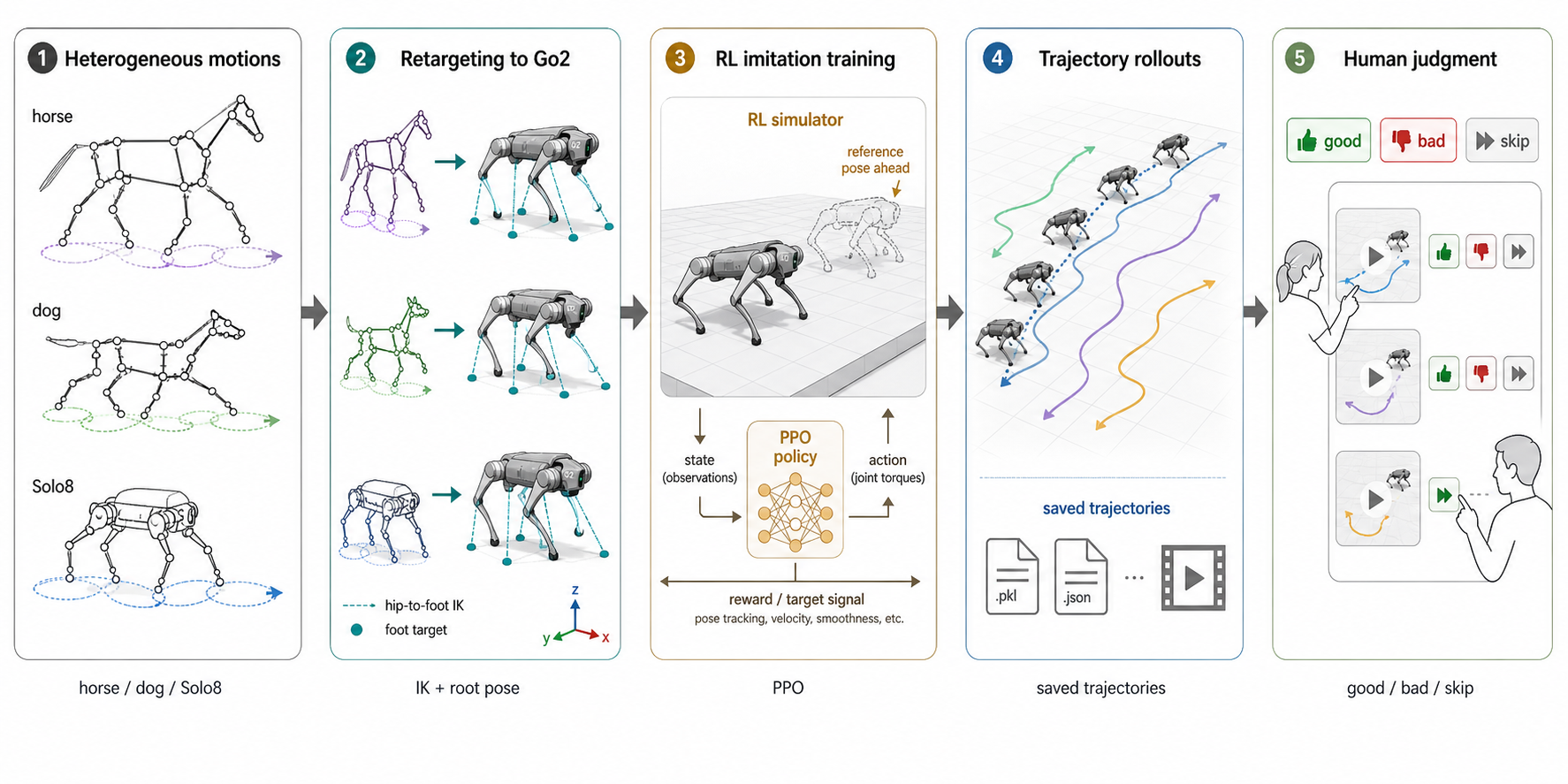}
  \end{center}
  \caption{Overview of the entire pipeline}
  \label{fig:pipeline}
\end{figure}

To easily gather the dataset of motions, we have created a pipeline consisting of 3 main stages:
\begin{enumerate}
    \item \textbf{Kinematic Retargeting} \\
    To leverage motion capture (MoCap) data from different source morphologies, we implement a retargeting procedure that maps source data onto the specific kinematic structure and degrees of freedom (DoF) of the Unitree Go2 robot. This ensures trajectory alignment while preserving the general motion present in the source.
    \item \textbf{Reinforcement Learning Motion Imitation} \\
    The retargeted motions serve as reference trajectories for an RL-based imitation framework. We optimize a distinct control policy for each reference motion, employing a reward function designed to minimize the deviation between the simulated robot's state and the reference trajectory, thus producing robust, physically feasible behaviors.
    \item \textbf{Trajectory gathering and filtering}\\
    The optimized policies are deployed in the simulator to generate a diverse set of stochastic rollouts. This process yields data containing both proprioceptive states and corresponding control actions (motor commands). To ensure dataset quality, we filter out unstable trajectories, such as those involving collisions, falls, or significant deviations from the reference motion.
\end{enumerate}
We describe details of each of the above steps in the following subsections and illustrate the entire pipeline in Figure \ref{fig:pipeline}. The pipeline follows the general outline of other similar efforts \citep{motionimitation, MimicKitPeng2025}. A key contribution of this work, beyond the dataset itself, is the implementation of this pipeline within Genesis \citep{Genesis}, a high-performance, GPU-accelerated physics engine, able to run on both consumer-grade and cluster-grade GPUs. Given that our approach necessitates training individual policies for a vast library of motions, the massive parallelism and computational efficiency offered by Genesis were critical.

\subsection{Kinematic Retargeting}

To leverage diverse quadrupedal datasets, we implement a kinematic retargeting pipeline designed to map source motions onto the specific mechanical constraints and degrees of freedom (DoF) of the Unitree Go2 platform. For each data source, we define a separate class that maps the morphology-specific data to full pose information for the Go2 robot. The retargeting process is executed frame-wise. For each timestep in the source trajectory, we extract the relevant target coordinates and resolve the corresponding joint configurations using an inverse kinematics (IK) solver optimized for the Go2's 12-DoF morphology. By treating the retargeted foot positions as task-space objectives, we ensure that the resulting poses preserve the essential gait characteristics of the source data while remaining strictly within the robot's reachable workspace. This modular design facilitates high scalability; incorporating novel data sources requires only the implementation of a specific mapping class. The output of this stage is a continuous, robot-compliant reference trajectory, which serves as the ground truth for the subsequent imitation learning phase. It is important to stress that the retargeted kinematic motions do \textbf{not} contain action data. They are stop-motion frames, with no information about the movement in between each pair of frames. Because of this, the data cannot be directly used for most training regimes, like behavioral cloning.

\subsection{Motion Imitation}
For each kinematic reference trajectory, we optimize a dedicated control policy using a Reinforcement Learning (RL) based imitation framework, following the methodology proposed by \citep{motionimitation}. The task is formulated as a goal-conditioned MDP where the objective is to minimize the tracking error between the simulated agent and the reference motion.

\begin{figure}
  \begin{center}
    \includegraphics[width=0.95\textwidth]{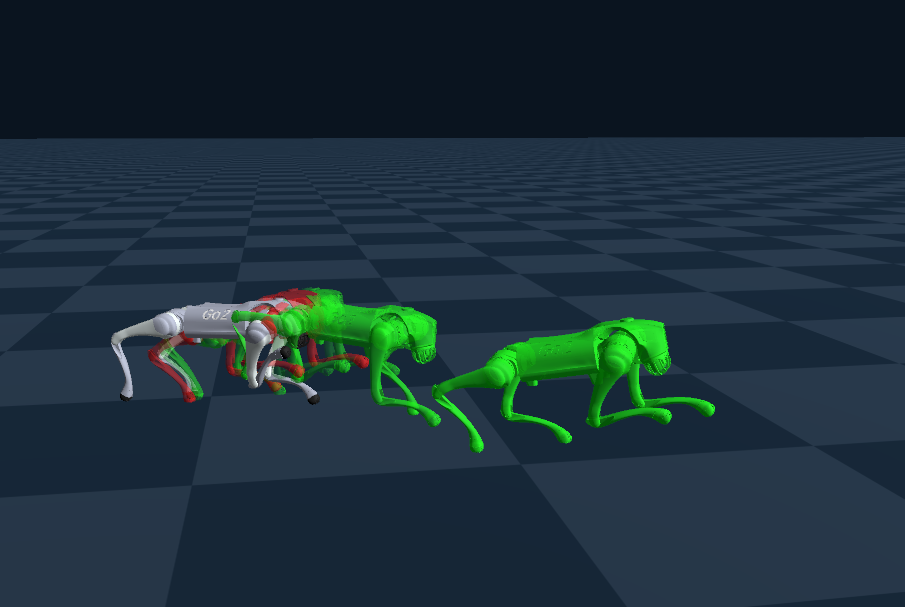}
  \end{center}
  \caption{Simulation and observation - The red contour of the robot represents the current state of the reference motion, while four green contours represent the state of the reference motion in 1,2, 10, and 30 timesteps, which are given to the policy.}
  \label{fig:observation}
\end{figure}

At the onset of each episode, the robot is initialized to the state corresponding to the first frame of the reference trajectory. To ensure heading invariance, a random heading rotation is applied to the global orientation of the reference motion at initialization. The observation space provided to the policy includes the robot's current proprioceptive state (e.g., joint positions, velocities, and base orientation) as well as reference trajectory state in the future. Specifically, we provide the target poses from the reference trajectory sampled at $t+1, t+2, t+10$, and $t+30$ time steps (see Figure \ref{fig:observation}).

The reward function is defined as a weighted sum of several tracking terms, penalizing deviations in root position, linear and angular velocities, and joint-space configurations. We utilize the Proximal Policy Optimization (PPO) algorithm \citep{ppo} as implemented in the RSL-RL library \citep{rslrl}. Leveraging the massive parallelism of the Genesis engine, we deploy 8,192 environments simultaneously. This high-throughput configuration enables each policy to be trained for approximately 2 billion environment steps over 10,000 iterations, ensuring convergence to a saturated and robust state. Since most motions are quite short, we apply periodic wrapping to the reference data, allowing for the generation of trajectories of arbitrary length.

\subsection{Trajectory gathering and filtering}

Following policy convergence, we execute stochastic rollouts to generate the trajectories of the \texttt{Kine2Go} dataset. To maximize the diversity of the covered state-space, each rollout is initialized with a randomized heading orientation. For a specified number of frames, we record the proprioceptive state and the corresponding control actions of the robot. For each trajectory, we archive a snapshot of the simulator's internal state to facilitate exact reproducibility within the Genesis engine. However, the primary trajectory data is stored in an engine-agnostic format to ensure cross-platform compatibility. The recorded state vector for each frame includes:
\begin{enumerate}
\item \textbf{Joint Configurations:} Local positions and velocities of the degrees of freedom (DoF) relative to the parent links.\item \textbf{Base Kinematics:} The orientation of the base link represented as a unit quaternion and its associated angular velocity.
\item \textbf{Control Actions:} The normalized motor commands (desired offsets w.r.t default positions) issued by the policy.
\item \textbf{Global States:} The absolute Cartesian positions and linear velocities of the joint centers in the world frame.
\end{enumerate}
Due to the inherent stochasticity of RL, the deployed policy can occasionally yield sub-optimal behaviors, such as dynamic instability or tracking divergence. To address this, each trajectory is rendered into a high-resolution video and manually inspected for physical plausibility and adherence to the reference motion. In case of trajectories that are harder to learn, for example involving complex loops, fast turning, or high speed, it may be necessary to deploy the policy for more episodes to compensate for this difficulty. Our pipeline provides integrated utility scripts to streamline this verification and filtering process. For full details regarding each trajectory, please refer to Appendix \ref{app:data}. To further increase the quality of the included data, we have trimmed the resulting trajectories by deleting the first 0.5 seconds, in order for the resulting trajectory not to include the starting "warmup" frames, where the policy tries to get in sync with the reference motion.

Thanks to fast physical simulation and the fact that we can produce a large number of distinct, although similar, motions from each policy, our pipeline doesn't need vast computational resources, and can be successfully used on both small GPU clusters and consumer-grade GPU workstations, thus democratizing the process of robotic data generation for quadrupedal robots.

\section{Data sources for \texttt{Kine2Go}} \label{sec:data}
To ensure diverse data, when creating our dataset, we used 4 distinct data sources from motion capture of different morphologies. Since during a single episode a reference motion is cycled multiple times, the first and last frames of the motion must have \textit{similar} joint positions, to ensure a smooth transition between cycles. To that end, we had to trim each reference motion, choosing the best start and end frame. This process was done manually by watching the videos of the motions retargeted to the Go2 robot. The complete table outlining all the data sources can be found in Appendix \ref{app:sources}.

\subsection{AI4Animation}
Since our methodology is modeled after \citep{motionimitation}, we also used some of the data present in that work. Specifically, we used the motion capture data, gathered from a dog, provided with the code repository. We have used 6 of the trimmed motions used in the original paper, as well as 9 other trimmed by us manually.
\subsection{VHDC}
This is a MoCap dataset gathered from horses \citep{vhdc2020}. For our research, we restricted our usage to the first dataset provided by this source, namely the Haflinger dataset setup. From this collection, we chose 6 "trot" and 6 "walk" motions originating from the MoCap of two different horses, trimming the trajectories to fit our requirements. Although the full dataset provides more motions, many are quite similar to one another, so we decided to only use a curated subset.
\subsection{Solo8}
Solo8 is another quadrupedal robot. Its morphology differs from Unitree Go2 in two significant ways: first, it has only 8 degrees of freedom; it has hip and elbow joints, but no shoulder joint, thus the legs move in a plane perpendicular to the torso plane. Secondly, its front and back legs have opposing orientations of elbows, in contrast to Go2, where both front and back elbows are oriented in the same way. An image of the Solo8 robot is presented in Figure \ref{fig:solo8morph} The dataset we used \citep{solo8data} also provides some motions that are harder to replicate on the Go2 robot, due to morphological differences, for example, walking while crouched close to the ground. As such, data from the Solo8 robot provides not only a source of diverse trajectories, hard to find elsewhere, but also a test for our pipeline.
\subsection{AI4Animation interactive demo}
\begin{figure}[t!]
    \centering
    \begin{subfigure}[t]{0.5\textwidth}
        \centering
        \includegraphics[width=0.95\textwidth]{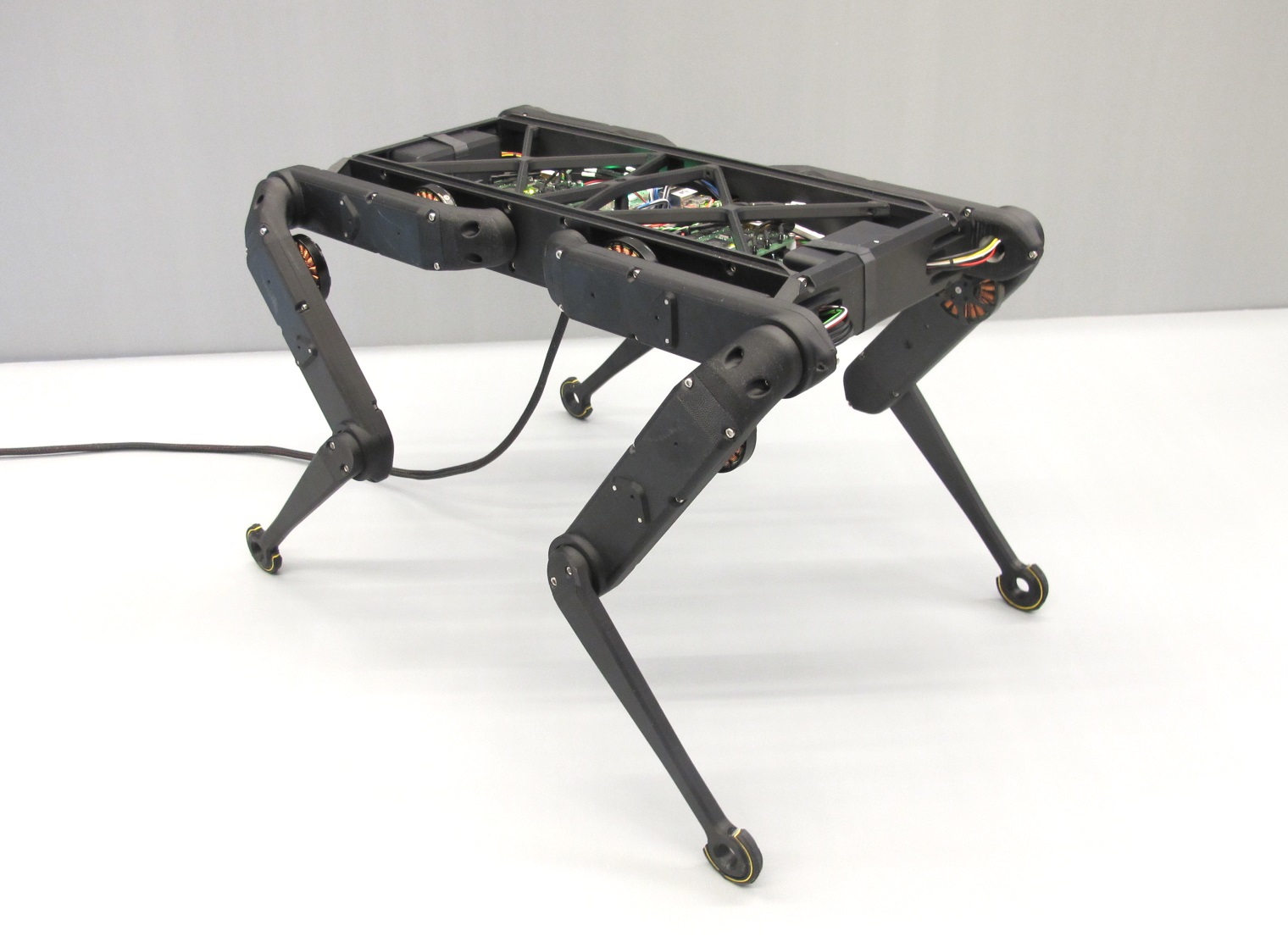}
        \caption{Solo8 robot has morphology distinctly different than Unitree Go2. Image sourced from \citep{solo8photo}. Licensed under CC BY 3.0. }
        \label{fig:solo8morph}
    \end{subfigure}%
    ~ 
    ~
    ~
    ~
    ~
    \begin{subfigure}[t]{0.5\textwidth}
        \centering
        \includegraphics[width=0.95\textwidth]{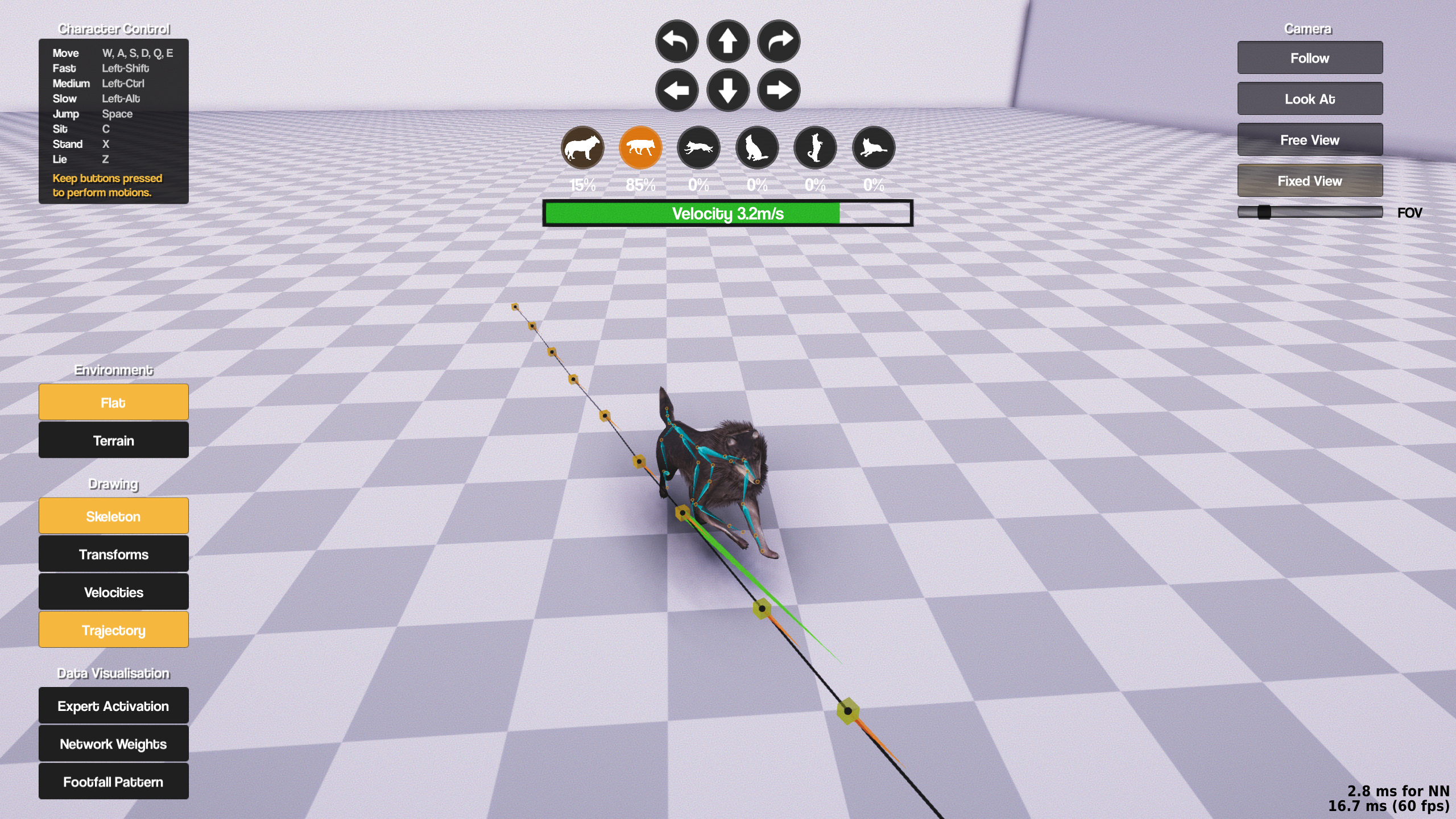}
        \caption{The wolf simulator from AI4 Animation. The user can issue commands to the model via keyboard, and thus, we can create complex trajectories.}
        \label{fig:wolfsim}
    \end{subfigure}
    \label{fig:data_sources}
    \caption{Solo8 robot and user-controlled wolf simulation.}
\end{figure}

We have also used an interactive simulation from \citep{DogInteractive}. The authors provided a Unity Engine-powered application, where users can steer an animated wolf (see Figure \ref{fig:wolfsim}). We modified the original application to capture the kinematic information of the animated dog, and then manually created 7 motions that include walking and running along more complex paths (i.e., ellipses, loops, squares, etc.).

\section{\texttt{Kine2Go} dataset} \label{sec:dataset}
Utilizing the pipeline described in Section \ref{sec:pipeline} and the data sources enumerated in Section \ref{sec:data}, we created \texttt{Kine2Go}: a comprehensive, high-fidelity dataset comprising 800 motion trajectories for the Unitree Go2 quadruped platform. The dataset is derived from 40 distinct Reinforcement Learning policies, each optimized for a unique reference motion. To maximize state-space coverage and diversity, each policy was deployed to generate 20 independent trajectories, characterized by randomized initial headings and inherent policy stochasticity. Depending on the type of motion, policies were deployed for between 5 and 20 seconds to capture at least one full cycle of the motion.

\begin{figure}[t!]
    \centering
    \begin{subfigure}[t]{0.5\textwidth}
        \centering
        \includegraphics[width=0.95\textwidth]{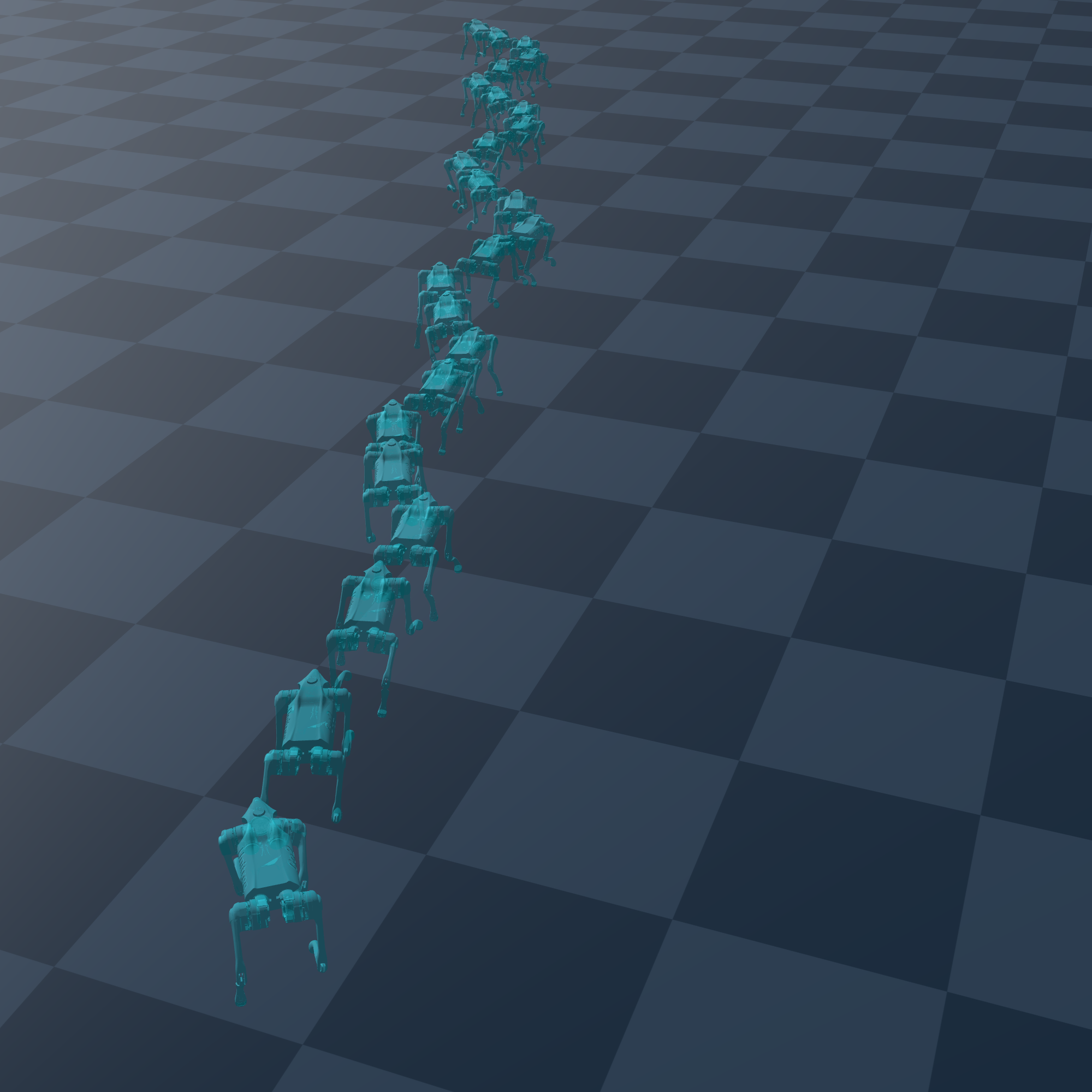}
        \caption{Visualization of a single trajectory \texttt{ai4\_dog\_synth\_wide\_strafe/traj\_0000}    }
        \label{fig:traj_vis}
    \end{subfigure}%
    ~ 
    ~
    ~
    ~
    ~
    \begin{subfigure}[t]{0.5\textwidth}
        \centering
        \includegraphics[width=0.95\textwidth]{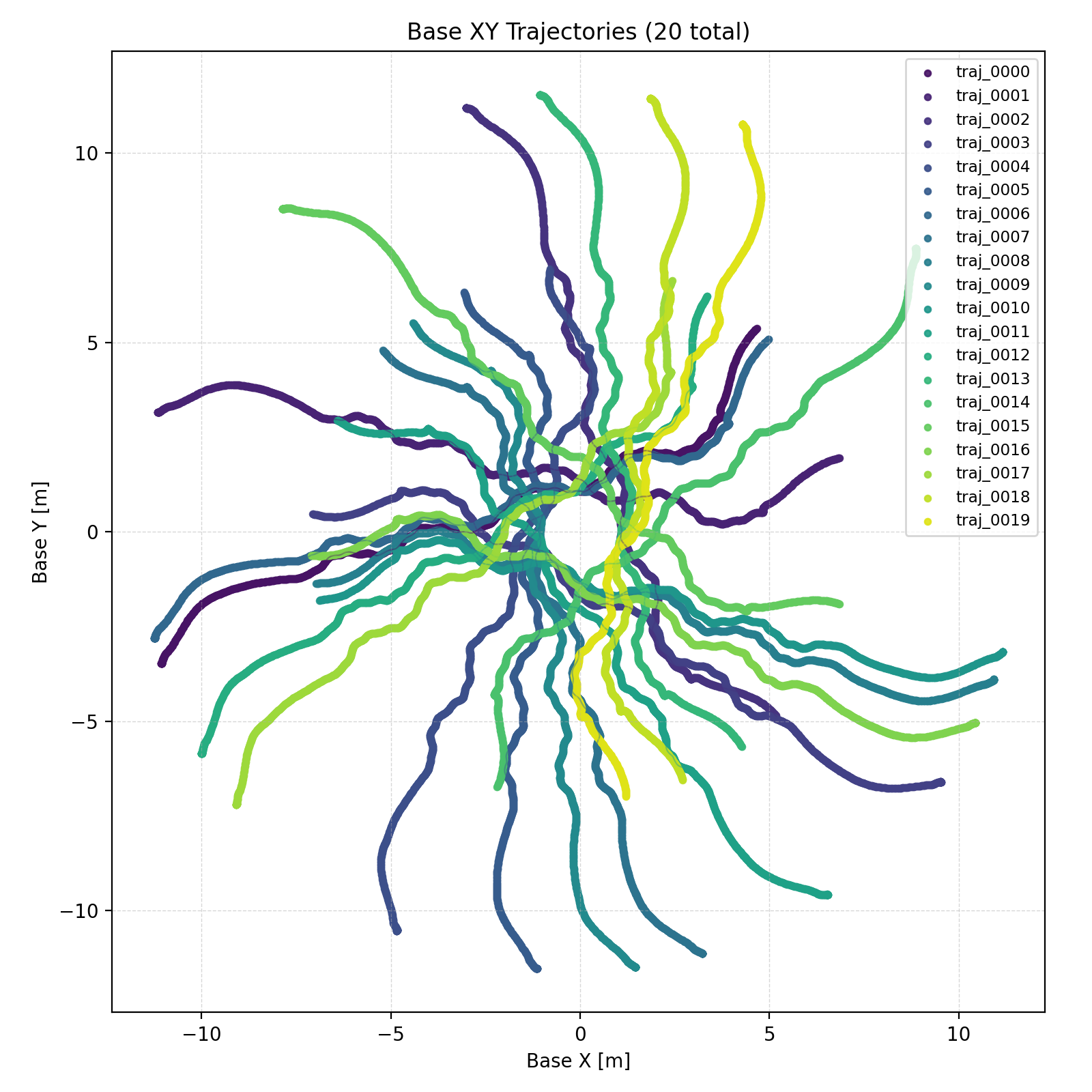}
        \caption{2D space coverage (position of the root link of the robot) of 20 trajectories generated from the same policy - \texttt{ai4\_dog\_synth\_wide\_strafe}}
        \label{fig:2d_cov}
    \end{subfigure}
    \caption{Visualization of example trajectory and motion space coverage}
\end{figure}

The resulting dataset contains a broad range of locomotive motions, including, but not limited to, walking, running, trotting, turning, and spinning. A detailed breakdown of these 40 motions, alongside a comprehensive description, is provided in Appendix \ref{app:data}.  We illustrate one trajectory in Figure \ref{fig:traj_vis}, while Figure \ref{fig:2d_cov} visualizes the example 2D planar coverage achieved across 20 trajectories for a singular motion.

Analogous to the utility of the AMASS or MoCapAct datasets for humanoid research, \texttt{Kine2Go} serves as an "off-the-shelf" solution for researchers working with the Unitree Go2 or similar quadrupedal morphologies. By providing an accessible alternative to complex humanoid embodiments, this dataset lowers the barrier to entry for advanced motion research. Consequently, it enables rapid iteration and experimentation, allowing for the seamless evaluation of novel methods in both simulation and real-world environments, thus pushing the frontier of agile quadrupedal robotics.

\subsection{Use cases and evaluation}
Due to its extensive state-space coverage and the inclusion of full kinematic data (including joint-level motor commands), \texttt{Kine2Go} is ideally suited for training foundational locomotion models, such as Meta Motivo \citep{metamotivo}, or for regularizing learned controllers to promote lifelike, stable, and natural motion. Our dataset is intended to promote work on a new generation of behavioral foundational models that can be first trained in simulation and regularized with \texttt{Kine2Go}, and eventually deployed to the real world to truly evaluate their capabilities and robustness.

The methodology of evaluating such models is still a subject of study. 
In the case of simpler behavioral cloning methods, which train a policy following one single motion from our dataset, it is sufficient to track simple metrics like mean square error between the joint positions in the reference motion and the motion rolled out from the policy.
However, in the case of more complex systems, especially behavioral foundational models (which we target as the main beneficiaries of our dataset), this is not enough. Foundational models are not meant to merely mimic the motions from the dataset, but rather clone their general characteristics, and as such, a simple MSE error is not enough. Moreover, depending on the exact input modality of the foundational model, the evaluation practice has to be adjusted in kind. In the Meta Motivo \citep{metamotivo} and its successor BFM-Zero \citep{bfmzero}, which to our knowledge are currently the only instances of a behavioral foundational model, the authors propose to use Earth Mover's Distance metric \citep{emd} to assess how closely the robot follows the motion that it is evaluated on. This metric captures the \textit{closeness} of two distributions, which better answers the question of whether the model followed a given trajectory.

Evaluating models operating on even richer modalities remains a problem for future work. For example, if behavioral foundational models accepted natural language instructions, it would be essential to assess whether a model instructed to "run in a given direction" was indeed running, not walking.

\section{Conclusion} \label{sec:conclusion}
We present \texttt{Kine2Go}, a novel and diverse kinematic state-action dataset for the Unitree Go2 robot, comprising 800 trajectories across 40 unique policies. By aggregating data from varied source morphologies and retargeting them onto a unified quadrupedal embodiment, we ensure a broad spectrum of motions that were previously unavailable for this platform. At the same time, the architecture of the pipeline enables easy extension of the used data sources for specific use cases and future work. While large-scale kinematic datasets have significantly advanced humanoid robotics, the quadrupedal domain has lacked a comparable, open-source resource that bridges the gap between raw motion and actionable RL policies.

The resulting dataset and accompanying pipeline are an aid to all researchers working on foundational locomotion methods for the Unitree Go2 robot, as well as all other quadrupedal robots. We hope that our contribution will help to discover, evaluate, and deploy novel algorithms leading to significant advancements across Robotics and applied Reinforcement Learning, both in quadrupedal locomotion and more broadly in foundational RL models.

\section{Discussion}
\subsection{Limitations} \label{sec:limit}
When creating \texttt{Kine2Go}, we strived to cover as much of the state space and feasible motions as possible. However, there are some motions we have not covered. The most notable category is motions involving jumps. The RL algorithm had trouble imitating motions involving jumps where not all the legs were on the ground. The quality of the produced trajectories was low, so we have decided not to include them. 

Another important category is trajectories that, in Motion Capture, involved a dog sitting. Due to morphological differences between dogs and the Unitree Go2 robot, retargeting a sitting dog results in the robot's elbow joints being below the ground level. To fix this, each sitting trajectory would have to be manually changed, frame by frame, to move each robot's foot to the correct position. It is also worth noting that all the trajectories gathered concern a flat terrain.

Lastly, the Unitree Go2 robot has pressure sensors in each foot. However, we have chosen not to simulate pressure sensors in our simulation, and as such, this data is not present in the dataset. 
\subsection{Future Work}
There are two main directions for future work. The first one is using our dataset to train foundational behavioral models and deploy them on the Unitree Go2 robot. This is one of the primary reasons we have created the dataset, and it is the main intended use of it. We believe that quadrupedal robotic platforms are excellent testing grounds for methods involving foundational behavioral models, because they are comparatively much simpler and more stable than humanoids. Because of that, the entry level and difficulties encountered are much lower, which encourages novel research and democratizes research opportunities.
The second main direction of future work is extending the dataset itself. The main missing parts are described in Section \ref{sec:limit}. The first one is adding acrobatic maneuvers, such as jumping or backflips, into the dataset. The second one is trying to reproduce the gaits and motions on rough, uneven terrain, or even in more complex environments, including ramps, stairs, etc.

\begin{ack}
This research was partially supported by National Science Centre, Poland, grant Sonata Bis 2024/54/E/ST6/00388.
\end{ack}

\bibliographystyle{plainnat}
\bibliography{ref.bib}

@article{DBLP:journals/corr/abs-1910-11215,
  author       = {Sudeep Dasari and
                  Frederik Ebert and
                  Stephen Tian and
                  Suraj Nair and
                  Bernadette Bucher and
                  Karl Schmeckpeper and
                  Siddharth Singh and
                  Sergey Levine and
                  Chelsea Finn},
  title        = {RoboNet: Large-Scale Multi-Robot Learning},
  journal      = {CoRR},
  volume       = {abs/1910.11215},
  year         = {2019},
  url          = {http://arxiv.org/abs/1910.11215},
  eprinttype   = {arXiv},
  eprint       = {1910.11215},
  bibsource    = {dblp computer science bibliography, https://dblp.org}
}

@misc{walke2024bridgedatav2datasetrobot,
      title={BridgeData V2: A Dataset for Robot Learning at Scale}, 
      author={Homer Walke and Kevin Black and Abraham Lee and Moo Jin Kim and Max Du and Chongyi Zheng and Tony Zhao and Philippe Hansen-Estruch and Quan Vuong and Andre He and Vivek Myers and Kuan Fang and Chelsea Finn and Sergey Levine},
      year={2024},
      eprint={2308.12952},
      archivePrefix={arXiv},
      primaryClass={cs.RO},
      url={https://arxiv.org/abs/2308.12952}, 
}

@misc{khazatsky2025droidlargescaleinthewildrobot,
      title={DROID: A Large-Scale In-The-Wild Robot Manipulation Dataset}, 
      author={Alexander Khazatsky and Karl Pertsch and Suraj Nair and Ashwin Balakrishna and Sudeep Dasari and Siddharth Karamcheti and Soroush Nasiriany and Mohan Kumar Srirama and Lawrence Yunliang Chen and Kirsty Ellis and Peter David Fagan and Joey Hejna and Masha Itkina and Marion Lepert and Yecheng Jason Ma and Patrick Tree Miller and Jimmy Wu and Suneel Belkhale and Shivin Dass and Huy Ha and Arhan Jain and Abraham Lee and Youngwoon Lee and Marius Memmel and Sungjae Park and Ilija Radosavovic and Kaiyuan Wang and Albert Zhan and Kevin Black and Cheng Chi and Kyle Beltran Hatch and Shan Lin and Jingpei Lu and Jean Mercat and Abdul Rehman and Pannag R Sanketi and Archit Sharma and Cody Simpson and Quan Vuong and Homer Rich Walke and Blake Wulfe and Ted Xiao and Jonathan Heewon Yang and Arefeh Yavary and Tony Z. Zhao and Christopher Agia and Rohan Baijal and Mateo Guaman Castro and Daphne Chen and Qiuyu Chen and Trinity Chung and Jaimyn Drake and Ethan Paul Foster and Jensen Gao and Vitor Guizilini and David Antonio Herrera and Minho Heo and Kyle Hsu and Jiaheng Hu and Muhammad Zubair Irshad and Donovon Jackson and Charlotte Le and Yunshuang Li and Kevin Lin and Roy Lin and Zehan Ma and Abhiram Maddukuri and Suvir Mirchandani and Daniel Morton and Tony Nguyen and Abigail O'Neill and Rosario Scalise and Derick Seale and Victor Son and Stephen Tian and Emi Tran and Andrew E. Wang and Yilin Wu and Annie Xie and Jingyun Yang and Patrick Yin and Yunchu Zhang and Osbert Bastani and Glen Berseth and Jeannette Bohg and Ken Goldberg and Abhinav Gupta and Abhishek Gupta and Dinesh Jayaraman and Joseph J Lim and Jitendra Malik and Roberto Martín-Martín and Subramanian Ramamoorthy and Dorsa Sadigh and Shuran Song and Jiajun Wu and Michael C. Yip and Yuke Zhu and Thomas Kollar and Sergey Levine and Chelsea Finn},
      year={2025},
      eprint={2403.12945},
      archivePrefix={arXiv},
      primaryClass={cs.RO},
      url={https://arxiv.org/abs/2403.12945}, 
}

@misc{embodimentcollaboration2025openxembodimentroboticlearning,
      title={Open X-Embodiment: Robotic Learning Datasets and RT-X Models}, 
      author={Embodiment Collaboration and Abby O'Neill and Abdul Rehman and Abhinav Gupta and Abhiram Maddukuri and Abhishek Gupta and Abhishek Padalkar and Abraham Lee and Acorn Pooley and Agrim Gupta and Ajay Mandlekar and Ajinkya Jain and Albert Tung and Alex Bewley and Alex Herzog and Alex Irpan and Alexander Khazatsky and Anant Rai and Anchit Gupta and Andrew Wang and Andrey Kolobov and Anikait Singh and Animesh Garg and Aniruddha Kembhavi and Annie Xie and Anthony Brohan and Antonin Raffin and Archit Sharma and Arefeh Yavary and Arhan Jain and Ashwin Balakrishna and Ayzaan Wahid and Ben Burgess-Limerick and Beomjoon Kim and Bernhard Schölkopf and Blake Wulfe and Brian Ichter and Cewu Lu and Charles Xu and Charlotte Le and Chelsea Finn and Chen Wang and Chenfeng Xu and Cheng Chi and Chenguang Huang and Christine Chan and Christopher Agia and Chuer Pan and Chuyuan Fu and Coline Devin and Danfei Xu and Daniel Morton and Danny Driess and Daphne Chen and Deepak Pathak and Dhruv Shah and Dieter Büchler and Dinesh Jayaraman and Dmitry Kalashnikov and Dorsa Sadigh and Edward Johns and Ethan Foster and Fangchen Liu and Federico Ceola and Fei Xia and Feiyu Zhao and Felipe Vieira Frujeri and Freek Stulp and Gaoyue Zhou and Gaurav S. Sukhatme and Gautam Salhotra and Ge Yan and Gilbert Feng and Giulio Schiavi and Glen Berseth and Gregory Kahn and Guangwen Yang and Guanzhi Wang and Hao Su and Hao-Shu Fang and Haochen Shi and Henghui Bao and Heni Ben Amor and Henrik I Christensen and Hiroki Furuta and Homanga Bharadhwaj and Homer Walke and Hongjie Fang and Huy Ha and Igor Mordatch and Ilija Radosavovic and Isabel Leal and Jacky Liang and Jad Abou-Chakra and Jaehyung Kim and Jaimyn Drake and Jan Peters and Jan Schneider and Jasmine Hsu and Jay Vakil and Jeannette Bohg and Jeffrey Bingham and Jeffrey Wu and Jensen Gao and Jiaheng Hu and Jiajun Wu and Jialin Wu and Jiankai Sun and Jianlan Luo and Jiayuan Gu and Jie Tan and Jihoon Oh and Jimmy Wu and Jingpei Lu and Jingyun Yang and Jitendra Malik and João Silvério and Joey Hejna and Jonathan Booher and Jonathan Tompson and Jonathan Yang and Jordi Salvador and Joseph J. Lim and Junhyek Han and Kaiyuan Wang and Kanishka Rao and Karl Pertsch and Karol Hausman and Keegan Go and Keerthana Gopalakrishnan and Ken Goldberg and Kendra Byrne and Kenneth Oslund and Kento Kawaharazuka and Kevin Black and Kevin Lin and Kevin Zhang and Kiana Ehsani and Kiran Lekkala and Kirsty Ellis and Krishan Rana and Krishnan Srinivasan and Kuan Fang and Kunal Pratap Singh and Kuo-Hao Zeng and Kyle Hatch and Kyle Hsu and Laurent Itti and Lawrence Yunliang Chen and Lerrel Pinto and Li Fei-Fei and Liam Tan and Linxi "Jim" Fan and Lionel Ott and Lisa Lee and Luca Weihs and Magnum Chen and Marion Lepert and Marius Memmel and Masayoshi Tomizuka and Masha Itkina and Mateo Guaman Castro and Max Spero and Maximilian Du and Michael Ahn and Michael C. Yip and Mingtong Zhang and Mingyu Ding and Minho Heo and Mohan Kumar Srirama and Mohit Sharma and Moo Jin Kim and Muhammad Zubair Irshad and Naoaki Kanazawa and Nicklas Hansen and Nicolas Heess and Nikhil J Joshi and Niko Suenderhauf and Ning Liu and Norman Di Palo and Nur Muhammad Mahi Shafiullah and Oier Mees and Oliver Kroemer and Osbert Bastani and Pannag R Sanketi and Patrick "Tree" Miller and Patrick Yin and Paul Wohlhart and Peng Xu and Peter David Fagan and Peter Mitrano and Pierre Sermanet and Pieter Abbeel and Priya Sundaresan and Qiuyu Chen and Quan Vuong and Rafael Rafailov and Ran Tian and Ria Doshi and Roberto Martín-Martín and Rohan Baijal and Rosario Scalise and Rose Hendrix and Roy Lin and Runjia Qian and Ruohan Zhang and Russell Mendonca and Rutav Shah and Ryan Hoque and Ryan Julian and Samuel Bustamante and Sean Kirmani and Sergey Levine and Shan Lin and Sherry Moore and Shikhar Bahl and Shivin Dass and Shubham Sonawani and Shubham Tulsiani and Shuran Song and Sichun Xu and Siddhant Haldar and Siddharth Karamcheti and Simeon Adebola and Simon Guist and Soroush Nasiriany and Stefan Schaal and Stefan Welker and Stephen Tian and Subramanian Ramamoorthy and Sudeep Dasari and Suneel Belkhale and Sungjae Park and Suraj Nair and Suvir Mirchandani and Takayuki Osa and Tanmay Gupta and Tatsuya Harada and Tatsuya Matsushima and Ted Xiao and Thomas Kollar and Tianhe Yu and Tianli Ding and Todor Davchev and Tony Z. Zhao and Travis Armstrong and Trevor Darrell and Trinity Chung and Vidhi Jain and Vikash Kumar and Vincent Vanhoucke and Vitor Guizilini and Wei Zhan and Wenxuan Zhou and Wolfram Burgard and Xi Chen and Xiangyu Chen and Xiaolong Wang and Xinghao Zhu and Xinyang Geng and Xiyuan Liu and Xu Liangwei and Xuanlin Li and Yansong Pang and Yao Lu and Yecheng Jason Ma and Yejin Kim and Yevgen Chebotar and Yifan Zhou and Yifeng Zhu and Yilin Wu and Ying Xu and Yixuan Wang and Yonatan Bisk and Yongqiang Dou and Yoonyoung Cho and Youngwoon Lee and Yuchen Cui and Yue Cao and Yueh-Hua Wu and Yujin Tang and Yuke Zhu and Yunchu Zhang and Yunfan Jiang and Yunshuang Li and Yunzhu Li and Yusuke Iwasawa and Yutaka Matsuo and Zehan Ma and Zhuo Xu and Zichen Jeff Cui and Zichen Zhang and Zipeng Fu and Zipeng Lin},
      year={2025},
      eprint={2310.08864},
      archivePrefix={arXiv},
      primaryClass={cs.RO},
      url={https://arxiv.org/abs/2310.08864}, 
}

@article{Peng_2018,
   title={DeepMimic: example-guided deep reinforcement learning of physics-based character skills},
   volume={37},
   ISSN={1557-7368},
   url={http://dx.doi.org/10.1145/3197517.3201311},
   DOI={10.1145/3197517.3201311},
   number={4},
   journal={ACM Transactions on Graphics},
   publisher={Association for Computing Machinery (ACM)},
   author={Peng, Xue Bin and Abbeel, Pieter and Levine, Sergey and van de Panne, Michiel},
   year={2018},
   pages={1–14} 
}

@misc{zhu2023h3wbhuman36m3dwholebody,
      title={H3WB: Human3.6M 3D WholeBody Dataset and Benchmark}, 
      author={Yue Zhu and Nermin Samet and David Picard},
      year={2023},
      eprint={2211.15692},
      archivePrefix={arXiv},
      primaryClass={cs.CV},
      url={https://arxiv.org/abs/2211.15692}, 
}

@misc{mahmood2019amassarchivemotioncapture,
      title={AMASS: Archive of Motion Capture as Surface Shapes}, 
      author={Naureen Mahmood and Nima Ghorbani and Nikolaus F. Troje and Gerard Pons-Moll and Michael J. Black},
      year={2019},
      eprint={1904.03278},
      archivePrefix={arXiv},
      primaryClass={cs.CV},
      url={https://arxiv.org/abs/1904.03278}, 
}

@misc{xu2024animal3dcomprehensivedataset3d,
      title={Animal3D: A Comprehensive Dataset of 3D Animal Pose and Shape}, 
      author={Jiacong Xu and Yi Zhang and Jiawei Peng and Wufei Ma and Artur Jesslen and Pengliang Ji and Qixin Hu and Jiehua Zhang and Qihao Liu and Jiahao Wang and Wei Ji and Chen Wang and Xiaoding Yuan and Prakhar Kaushik and Guofeng Zhang and Jie Liu and Yushan Xie and Yawen Cui and Alan Yuille and Adam Kortylewski},
      year={2024},
      eprint={2308.11737},
      archivePrefix={arXiv},
      primaryClass={cs.CV},
      url={https://arxiv.org/abs/2308.11737}, 
}

@article{li2024poses,
        title={The Poses for Equine Research Dataset (PFERD)},
        author={Li, Ci and Mellbin, Ylva and Krogager, Johanna and Polikovsky, Senya and Holmberg, Martin and Ghorbani, Nima and Black, Michael J and Kjellstr{\"o}m, Hedvig and Zuffi, Silvia and Hernlund, Elin},
        journal={Scientific Data},
        volume={11},
        number={1},
        pages={497},
        year={2024},
        publisher={Nature Publishing Group UK London}
      }

@inproceedings{
yu2021visuallocomotion,
title={Visual-Locomotion: Learning to Walk on Complex Terrains with Vision},
author={Wenhao Yu and Deepali Jain and Alejandro Escontrela and Atil Iscen and Peng Xu and Erwin Coumans and Sehoon Ha and Jie Tan and Tingnan Zhang},
booktitle={5th Annual Conference on Robot Learning },
year={2021},
url={https://openreview.net/forum?id=NDYbXf-DvwZ}
}

@misc{wagener2023mocapactmultitaskdatasetsimulated,
      title={MoCapAct: A Multi-Task Dataset for Simulated Humanoid Control}, 
      author={Nolan Wagener and Andrey Kolobov and Felipe Vieira Frujeri and Ricky Loynd and Ching-An Cheng and Matthew Hausknecht},
      year={2023},
      eprint={2208.07363},
      archivePrefix={arXiv},
      primaryClass={cs.RO},
      url={https://arxiv.org/abs/2208.07363}, 
}

@article{
   Genesis,
   author = {{Genesis AI Team}},
   title = {The Role of Simulation in Scalable Robotics, Genesis World 1.0, and the Path Forward},
   journal = {Genesis AI Blog},
   month = {May},
   year = {2026},
   url = {https://www.genesis.ai/blog/the-role-of-simulation-in-scalable-robotics-genesis-world-10-and-the-path-forward},
}

@article{motionimitation,
  title={Learning agile robotic locomotion skills by imitating animals},
  author={Peng, Xue Bin and Coumans, Erwin and Zhang, Tingnan and Lee, Tsang-Wei and Tan, Jie and Levine, Sergey},
  journal={arXiv preprint arXiv:2004.00784},
  year={2020}
}

@article{ppo,
  title={Proximal policy optimization algorithms},
  author={Schulman, John and Wolski, Filip and Dhariwal, Prafulla and Radford, Alec and Klimov, Oleg},
  journal={arXiv preprint arXiv:1707.06347},
  year={2017}
}

@article{rslrl,
  title={RSL-RL: A Learning Library for Robotics Research},
  author={Schwarke, Clemens and Mittal, Mayank and Rudin, Nikita and Hoeller, David and Hutter, Marco},
  journal={arXiv preprint arXiv:2509.10771},
  year={2025}
}

@article{MimicKitPeng2025,
      title={MimicKit: A Reinforcement Learning Framework for Motion Imitation and Control}, 
      author={Peng, Xue Bin},
      year={2025},
      journal={arXiv preprint arXiv:2510.13794},
      eprint={2510.13794},
      archivePrefix={arXiv},
      primaryClass={cs.GR},
      url={https://arxiv.org/abs/2510.13794}, 
}

@article{DogInteractive,
    author = {Zhang, He and Starke, Sebastian and Komura, Taku and Saito, Jun},
    title = {Mode-adaptive neural networks for quadruped motion control},
    year = {2018},
    issue_date = {August 2018},
    publisher = {Association for Computing Machinery},
    address = {New York, NY, USA},
    volume = {37},
    number = {4},
    issn = {0730-0301},
    url = {https://doi.org/10.1145/3197517.3201366},
    doi = {10.1145/3197517.3201366},
    journal = {ACM Trans. Graph.},
    month = jul,
    articleno = {145},
    numpages = {11}
}

@inproceedings{solo8data,
  title={Versatile skill control via self-supervised adversarial imitation of unlabeled mixed motions},
  author={Li, Chenhao and Blaes, Sebastian and Kolev, Pavel and Vlastelica, Marin and Frey, Jonas and Martius, Georg},
  booktitle={2023 IEEE international conference on robotics and automation (ICRA)},
  pages={2944--2950},
  year={2023},
  organization={IEEE}
}

@article{solo8photo,
doi = {10.1088/1757-899X/1292/1/012004},
url = {https://doi.org/10.1088/1757-899X/1292/1/012004},
year = {2023},
month = {oct},
publisher = {IOP Publishing},
volume = {1292},
number = {1},
pages = {012004},
author = {Chandiramani, Vijay and Conn, Andrew T. and Hauser, Helmut},
title = {Quantifying embodiment towards building better robots based on muscle-driven models},
journal = {IOP Conference Series: Materials Science and Engineering}
}

@misc{vhdc2020,
  author       = {{University of Bonn}},
  title        = {Vienna Horse Data Collection (VHDC) - Horse Motion Capture Data},
  howpublished = {\url{https://horse.cs.uni-bonn.de/vhdc-home.html}},
  year         = {2020},
  note         = {Accessed: 2026-05-06}
}

@misc{metamotivo,
      title={Zero-Shot Whole-Body Humanoid Control via Behavioral Foundation Models}, 
      author={Andrea Tirinzoni and Ahmed Touati and Jesse Farebrother and Mateusz Guzek and Anssi Kanervisto and Yingchen Xu and Alessandro Lazaric and Matteo Pirotta},
      year={2025},
      eprint={2504.11054},
      archivePrefix={arXiv},
      primaryClass={cs.LG},
      url={https://arxiv.org/abs/2504.11054}, 
}

@misc{bfmzero,
      title={BFM-Zero: A Promptable Behavioral Foundation Model for Humanoid Control Using Unsupervised Reinforcement Learning}, 
      author={Yitang Li and Zhengyi Luo and Tonghe Zhang and Cunxi Dai and Anssi Kanervisto and Andrea Tirinzoni and Haoyang Weng and Kris Kitani and Mateusz Guzek and Ahmed Touati and Alessandro Lazaric and Matteo Pirotta and Guanya Shi},
      year={2025},
      eprint={2511.04131},
      archivePrefix={arXiv},
      primaryClass={cs.RO},
      url={https://arxiv.org/abs/2511.04131}, 
}

@misc{nvidiasonic,
      title={SONIC: Supersizing Motion Tracking for Natural Humanoid Whole-Body Control}, 
      author={Zhengyi Luo and Ye Yuan and Tingwu Wang and Chenran Li and Sirui Chen and Fernando Castañeda and Zi-Ang Cao and Jiefeng Li and David Minor and Qingwei Ben and Xingye Da and Runyu Ding and Cyrus Hogg and Lina Song and Edy Lim and Eugene Jeong and Tairan He and Haoru Xue and Wenli Xiao and Zi Wang and Simon Yuen and Jan Kautz and Yan Chang and Umar Iqbal and Linxi "Jim" Fan and Yuke Zhu},
      year={2025},
      eprint={2511.07820},
      archivePrefix={arXiv},
      primaryClass={cs.RO},
      url={https://arxiv.org/abs/2511.07820}, 
}

@misc{yang2025generalizedanimalimitatoragile,
      title={Generalized Animal Imitator: Agile Locomotion with Versatile Motion Prior}, 
      author={Ruihan Yang and Zhuoqun Chen and Jianhan Ma and Chongyi Zheng and Yiyu Chen and Quan Nguyen and Xiaolong Wang},
      year={2025},
      eprint={2310.01408},
      archivePrefix={arXiv},
      primaryClass={cs.RO},
      url={https://arxiv.org/abs/2310.01408}, 
}

@article{emd,
    author = {Rubner, Yossi and Tomasi, Carlo and Guibas, Leonidas},
    year = {2000},
    month = {11},
    pages = {99-121},
    title = {The Earth Mover's Distance as a Metric for Image Retrieval},
    volume = {40},
    journal = {International Journal of Computer Vision},
    doi = {10.1023/A:1026543900054}
}

@misc{reward_hacking,
    title = {Faulty reward functions in the wild},
    url = {https://openai.com/index/faulty-reward-functions/},
    author = {Jack Clark and Dario Amodei},
    year = {2016},
}


\appendix

\section{Data sources} \label{app:sources}
Table \ref{tab:ranges} details the original reference motions used for retargeting across the 40 policies. The file identifiers correspond to the directory structure in the \texttt{motion\_retargeting/data/} folder of our provided code. Notably, \texttt{half\_flip\_jump} utilizes the same reference motion as \texttt{ellipse\_walk}. The resulting jumping behavior was an artifact of the policy's inability to accurately track the reference; however, we chose to include this emergent motion in the dataset.

\begin{table}[h]
\centering
\small
\setlength{\tabcolsep}{4pt}
\begin{tabular}{llcc}
\hline
\textbf{Policy} & \textbf{Reference File} & \textbf{Start frame} & \textbf{End frame} \\ \hline
\texttt{ai4\_dog\_canter} & \texttt{dog\_run00\_joint\_pos.txt} & 430 & 459 \\
\texttt{ai4\_dog\_left\_turn} & \texttt{dog\_walk09\_joint\_pos.txt} & 1085 & 1124 \\
\texttt{ai4\_dog\_pace} & \texttt{dog\_walk00\_joint\_pos.txt} & 162 & 201 \\
\texttt{ai4\_dog\_right\_turn} & \texttt{dog\_walk09\_joint\_pos.txt} & 2404 & 2450 \\
\texttt{ai4\_dog\_run\_00} & \texttt{dog\_run00\_joint\_pos.txt} & 399 & 535 \\
\texttt{ai4\_dog\_run\_02} & \texttt{dog\_run02\_joint\_pos.txt} & 35 & 196 \\
\texttt{ai4\_dog\_run\_04} & \texttt{dog\_run04\_joint\_pos.txt} & 493 & 716 \\
\texttt{ai4\_dog\_synth\_circle\_walk} & \texttt{circle\_walk.txt} & 446 & 806 \\
\texttt{ai4\_dog\_synth\_eight\_walk} & \texttt{eight\_walk.txt} & 0 & -1 \\
\texttt{ai4\_dog\_synth\_ellipse\_walk} & \texttt{ellipse\_walk.txt} & 116 & 1165 \\
\texttt{ai4\_dog\_synth\_half\_flip\_jump} & \texttt{ellipse\_walk.txt} & 116 & 1165 \\
\texttt{ai4\_dog\_synth\_square\_walk} & \texttt{square\_walk.txt} & 29 & 1087 \\
\texttt{ai4\_dog\_synth\_tight\_strafe} & \texttt{tight\_strafe.txt} & 54 & 962 \\
\texttt{ai4\_dog\_synth\_wide\_strafe} & \texttt{wide\_strafe.txt} & 121 & 1080 \\
\texttt{ai4\_dog\_trot\_00} & \texttt{dog\_walk03\_joint\_pos.txt} & 448 & 481 \\
\texttt{ai4\_dog\_trot\_01} & \texttt{dog\_run04\_joint\_pos.txt} & 630 & 663 \\
\texttt{ai4\_dog\_walk\_00} & \texttt{dog\_walk00\_joint\_pos.txt} & 101 & 580 \\
\texttt{ai4\_dog\_walk\_01} & \texttt{dog\_walk01\_joint\_pos.txt} & 377 & 1050 \\
\texttt{ai4\_dog\_walk\_02} & \texttt{dog\_walk02\_joint\_pos.txt} & 429 & 879 \\
\texttt{ai4\_dog\_walk\_03} & \texttt{dog\_walk03\_joint\_pos.txt} & 177 & 545 \\
\texttt{ai4\_dog\_walk\_04} & \texttt{dog\_walk04\_joint\_pos.txt} & 234 & 533 \\
\texttt{ai4\_dog\_walk\_06} & \texttt{dog\_walk06\_joint\_pos.txt} & 200 & 392 \\
\texttt{solo8\_crawl\_fast} & \texttt{solo8\_motion\_data.pt:3} & 77 & 119 \\
\texttt{solo8\_crawl\_slow} & \texttt{solo8\_motion\_data.pt:0} & 25 & -1 \\
\texttt{solo8\_jump\_forward\_a} & \texttt{solo8\_motion\_data.pt:23} & 52 & 114 \\
\texttt{solo8\_jump\_forward\_b} & \texttt{solo8\_motion\_data.pt:26} & 37 & 112 \\
\texttt{solo8\_scoot\_forward} & \texttt{solo8\_motion\_data.pt:1} & 69 & 119 \\
\texttt{solo8\_walk} & \texttt{solo8\_motion\_data.pt:2} & 49 & 84 \\
\texttt{vhdc\_horse1\_s1\_trot\_01} & \texttt{Horse1\_M1\_trot1\_kinematics.csv} & 13 & 594 \\
\texttt{vhdc\_horse1\_s1\_trot\_02} & \texttt{Horse1\_M1\_trot2\_kinematics.csv} & 2 & 599 \\
\texttt{vhdc\_horse1\_s1\_trot\_03} & \texttt{Horse1\_M1\_trot3\_kinematics.csv} & 12 & 567 \\
\texttt{vhdc\_horse1\_s1\_walk\_01} & \texttt{Horse1\_M1\_walk1\_kinematics.csv} & 9 & 599 \\
\texttt{vhdc\_horse1\_s1\_walk\_02} & \texttt{Horse1\_M1\_walk2\_kinematics.csv} & 4 & 591 \\
\texttt{vhdc\_horse1\_s1\_walk\_03} & \texttt{Horse1\_M1\_walk3\_kinematics.csv} & 9 & 598 \\
\texttt{vhdc\_horse1\_s2\_trot\_01} & \texttt{Horse1\_M2\_trot1\_kinematics.csv} & 7 & 1144 \\
\texttt{vhdc\_horse1\_s2\_trot\_02} & \texttt{Horse1\_M2\_trot2\_kinematics.csv} & 3 & 1173 \\
\texttt{vhdc\_horse1\_s2\_trot\_03} & \texttt{Horse1\_M2\_trot3\_kinematics.csv} & 10 & 1194 \\
\texttt{vhdc\_horse1\_s2\_walk\_01} & \texttt{Horse1\_M2\_walk1\_kinematics.csv} & 16 & 1145 \\
\texttt{vhdc\_horse1\_s2\_walk\_02} & \texttt{Horse1\_M2\_walk2\_kinematics.csv} & 27 & 1153 \\
\texttt{vhdc\_horse1\_s2\_walk\_03} & \texttt{Horse1\_M2\_walk3\_kinematics.csv} & 40 & 1137 \\ \hline \\
\end{tabular}
\caption{Frame ranges used for reference motion in each policy training, -1 means the last frame.}
\label{tab:ranges}
\end{table}

\section{Motion data details} \label{app:data}
Table \ref{tab:data} provides a comprehensive list of all motions, including brief descriptions for those that are not self-explanatory. We also specify the rollout lengths for the 20 trajectories sampled for each motion.

\begin{table}[t]
\centering
\begin{tabular}{llcc}
\hline
Name & Description & Length & Sample \\
\hline
    \texttt{ai4\_dog\_canter} & Cantering (running) gait & 571 & \checkmark \\
    \texttt{ai4\_dog\_left\_turn} & Turning left & 571 & \\
    \texttt{ai4\_dog\_pace} & Pacing gait & 571 & \\
    \texttt{ai4\_dog\_right\_turn} & Turning right & 571 & \checkmark \\
    \texttt{ai4\_dog\_run\_00} &  & 571 & \\
    \texttt{ai4\_dog\_run\_02} &  & 571 & \\
    \texttt{ai4\_dog\_run\_04} &  & 571 & \\
    \texttt{ai4\_dog\_synth\_circle\_walk} & Circular walking path & 1171 &  \checkmark \\
    \texttt{ai4\_dog\_synth\_eight\_walk} & Figure-eight walking path & 1171 & \\
    \texttt{ai4\_dog\_synth\_ellipse\_walk} & Elliptical walking path & 1171 & \\
    \texttt{ai4\_dog\_synth\_half\_flip\_jump} & Horizontal 180\textdegree jump & 300 & \checkmark\\
    \texttt{ai4\_dog\_synth\_square\_walk} & Square walking path & 1171 & \\
    \texttt{ai4\_dog\_synth\_tight\_strafe} & Tight strafing motion & 1171 & \\
    \texttt{ai4\_dog\_synth\_wide\_strafe} & Wide strafing motion & 960 & \\
    \texttt{ai4\_dog\_trot\_00} &  & 571 & \\
    \texttt{ai4\_dog\_trot\_01} &  & 571 & \\
    \texttt{ai4\_dog\_walk\_00} &  & 571 & \\
    \texttt{ai4\_dog\_walk\_01} &  & 571 & \\
    \texttt{ai4\_dog\_walk\_02} &  & 571 & \\
    \texttt{ai4\_dog\_walk\_03} &  & 571 & \\
    \texttt{ai4\_dog\_walk\_04} &  & 571 & \\
    \texttt{ai4\_dog\_walk\_06} &  & 571 & \\
    \texttt{solo8\_crawl\_fast} & Crawling close to the ground & 571 & \checkmark \\
    \texttt{solo8\_crawl\_slow} & Crawling very close to the ground & 571 & \\
    \texttt{solo8\_jump\_forward\_a} & Forward jump variant A & 571 & \checkmark \\
    \texttt{solo8\_jump\_forward\_b} & Forward jump variant B & 571 & \\
    \texttt{solo8\_scoot\_forward} & Forward scooting motion & 571 & \\
    \texttt{solo8\_walk} & Walking gait & 571 & \\
    \texttt{vhdc\_horse1\_s1\_trot\_01} &  & 571 & \checkmark \\
    \texttt{vhdc\_horse1\_s1\_trot\_02} &  & 571 & \\
    \texttt{vhdc\_horse1\_s1\_trot\_03} &  & 571 & \\
    \texttt{vhdc\_horse1\_s1\_walk\_01} &  & 571 & \checkmark \\
    \texttt{vhdc\_horse1\_s1\_walk\_02} &  & 571 & \\
    \texttt{vhdc\_horse1\_s1\_walk\_03} &  & 571 & \\
    \texttt{vhdc\_horse1\_s2\_trot\_01} &  & 571 & \\
    \texttt{vhdc\_horse1\_s2\_trot\_02} &  & 571 & \\
    \texttt{vhdc\_horse1\_s2\_trot\_03} &  & 571 & \\
    \texttt{vhdc\_horse1\_s2\_walk\_01} &  & 571 & \\
    \texttt{vhdc\_horse1\_s2\_walk\_02} &  & 571 & \\
    \texttt{vhdc\_horse1\_s2\_walk\_03} &  & 571 & \\
\hline \\
\end{tabular}

\caption{Motion dataset summary. The length is the number of frames in each of the 20 sampled trajectories.}
\label{tab:data}
\end{table}

\subsection{Sample selection} \label{app:sample}
The complete dataset, including videos, is approximately 30 GB in size. To facilitate the review process, we have provided a smaller sample dataset. This subset was explicitly chosen to represent a diverse range of behaviors, comprising a total of eight motions (two drawn from each of the four source morphologies). For each motion, we included two trajectories sampled from an RL policy trained on that specific task. The motions selected for this sample are denoted in Table \ref{tab:data}.

\section{Reinforcement Learning details} \label{app:rl}
For the motion imitation component of our pipeline, we reproduced the approach of \citep{motionimitation} within the Genesis simulation engine. Genesis supports GPU-vectorized, parallel environments, leading to substantially accelerated training times. We utilized the PPO algorithm \citep{ppo} from the RSL-RL library \citep{rslrl}, with specific hyperparameters detailed in Table \ref{tab:ppo}. Each policy was trained for 10,000 iterations. Because reference motions are typically shorter than a full episode, we implemented reference motion cycling, wherein a motion is repeated to fill the episode duration. This requires the start and end frames to be consistent in joint space, although the robot’s global position and orientation may vary. Policies were trained on NVIDIA A100 GPUs, with an average training time of 3 hours per motion. Subsequent trajectory sampling and video rendering were conducted on a standard, consumer-grade workstation without GPU acceleration.

\subsection{Reward structure}
The policy is trained using a weighted imitation reward designed to minimize the discrepancy between the agent's state and a time-aligned reference motion. At each control step $t$, the environment computes a reference state $\bar{s}_t$ comprising the root pose, root velocity, joint configurations, joint velocities, and end-effector positions. Following the reward structure described in \citep{motionimitation}, the total reward $R_t$ is defined as:

\begin{equation}
R_t
=
w_q r_q
+ w_{\dot{q}} r_{\dot{q}}
+ w_e r_e
+ w_p r_p
+ w_v r_v
\end{equation}

where the reward weights are:

\begin{equation}
w_q = 0.5,\quad
w_{\dot{q}} = 0.05,\quad
w_e = 0.2,\quad
w_p = 0.15,\quad
w_v = 0.1.
\end{equation}

Here $q_t$ and $\bar{q}_t$ denote the robot and reference joint angles, and
$\dot{q}_t$ and $\dot{\bar{q}}_t$ denote the corresponding joint velocities.

\begin{equation}
r_q =
\exp\left(
-5 \left\| q_t - \bar{q}_t \right\|_2^2
\right),
\end{equation}

\begin{equation}
r_{\dot{q}} =
\exp\left(
-0.1 \left\| \dot{q}_t - \dot{\bar{q}}_t \right\|_2^2
\right).
\end{equation}

The end-effector reward tracks the four feet relative to the robot root. Let
$f_{t,k}$ and $\bar{f}_{t,k}$ be the relative foot placement for foot $k$ at timestep $t$. The end-effector reward is:

\begin{equation}
r_e =
\exp\left(
-40 \sum_{k=1}^{4}
\left\| f_{t,k} - \bar{f}_{t,k}\right\|_2^2
\right).
\end{equation}

The root pose reward consists of separate position and orientation tracking terms.
Let $p_t$ and $\bar{p}_t$ be root positions, and let $Q_t$ and $\bar{Q}_t$ be root
orientations represented as quaternions. The orientation error is computed from the
axis-angle norm of the relative quaternion:

\begin{equation}
Q_{\mathrm{err}} = Q_t \bar{Q}_t^{-1},
\qquad
\theta_t = \left\|  \text{AxisAngle} (Q_{\mathrm{err}}) \right\|_2.
\end{equation}

The root pose reward is

\begin{equation}
r_p =
\exp\left(
-20 \left\| p_t - \bar{p}_t \right\|_2^2
\right)
+
\exp\left(
-10 \theta_t^2
\right).
\end{equation}

Finally, the root velocity reward tracks both linear and angular root velocities
in the world frame. Let $v_t, \omega_t$ be the robot root linear and angular
velocities, and $\bar{v}_t, \bar{\omega}_t$ the corresponding reference values:

\begin{equation}
r_v =
\exp\left(
-2 \left\| v_t - \bar{v}_t \right\|_2^2
\right)
+
\exp\left(
-0.2 \left\| \omega_t - \bar{\omega}_t \right\|_2^2
\right).
\end{equation}

\begin{table}[h]
\centering
\begin{tabular}{lll}
\hline
Category & Parameter & Value \\
\hline
Algorithm & Actor hidden dimensions & [512, 256, 128] \\
Algorithm & Critic hidden dimensions & [512, 256, 128] \\
Algorithm & Activation & ELU \\
Algorithm & Initial action noise std. & 1.0 \\
\hline
PPO optimization & Learning rate & 0.001 \\
PPO optimization & Learning rate schedule & adaptive \\
PPO optimization & PPO clip parameter & 0.2 \\
PPO optimization & Desired KL & 0.01 \\
PPO optimization & Entropy coefficient & 0.01 \\
PPO optimization & Value loss coefficient & 1.0 \\
PPO optimization & Use clipped value loss & true \\
PPO optimization & Max gradient norm & 1.0 \\
PPO optimization & Learning epochs per update & 5 \\
PPO optimization & Mini-batches per update & 4 \\
PPO optimization & Number of steps & 10000 \\
\hline
Returns & Discount factor $\gamma$ & 0.99 \\
Returns & GAE parameter $\lambda$ & 0.95 \\
\hline
Rollout & Steps per environment & 24 \\
Rollout & Maximum iterations & 10000 \\
Rollout & Save interval & 100 \\
Rollout & Seed & 1 \\
\hline
Environment & Number of parallel environments & 8192 \\
Environment & Episode length & 20.0 s \\
Environment & Control frequency & 60 Hz \\
Environment & Decimation & 4 \\
Environment & Action scale & 0.25 \\
Environment & Action latency & 0.02 s \\
Environment & Action clipping & 100.0 \\
Environment & Policy steps per reference motion step & 1 \\
\hline
Observation & History length & 3 \\
Observation & Observation noise & 0.0 \\
Observation & Empirical normalization & none \\
\hline
Imitation reward & Joint rotation scale & 0.008333 \\
Imitation reward & Joint velocity scale & 0.000833 \\
Imitation reward & End-effector position scale & 0.003333 \\
Imitation reward & Root pose scale & 0.0025 \\
Imitation reward & Root velocity scale & 0.001667 \\
\hline
Initialization & Reference-state init probability & 0.9 \\
Initialization & Perturb initial state & true \\
Initialization & Joint position randomization range & 0.3 \\
Initialization & Base position randomization range & 1.0 \\
Initialization & Randomize base yaw & true \\
Initialization & Randomize trajectory heading & true \\
\hline
Termination & Distance threshold & 1.0 m \\
Termination & Rotation threshold & 1.5708 rad \\
\hline \\
\end{tabular}
\caption{PPO training hyperparameters used for motion imitation.}
\label{tab:ppo}
\end{table}


\newpage

\end{document}